\documentclass[runningheads,a4paper]{llncs}

\def\etal{\textit{et al.}\xspace}

\usepackage{xspace}
\usepackage{amssymb}
\usepackage{graphicx}
\usepackage{amsmath}

\usepackage{algorithm}
\usepackage[noend]{algpseudocode}
\usepackage{algorithmicx}
\usepackage{array}
\usepackage{color}

\usepackage{url}
\begin{document}

\mainmatter  % start of an individual contribution

% first the title is needed

\title{A Multi-Armed Bandit to Smartly Select a Training Set from Big Medical Data}
\titlerunning{A MAB to Smartly Select a Training Set from Big Medical Data} 

% the name(s) of the author(s) follow(s) next
%

% their surnames. This ensures that the names appear correctly in
% the running heads and the author index.
%
\author{Authors}
\author{Benjam\'in Guti\'errez \inst{1,2} \and
	  Lo\"ic Peter \inst{3}  \and 
	   Tassilo Klein \inst{4}  \and
	    Christian Wachinger \inst{1} 
	    \thanks{The final publication is available on the MICCAI 17 proceedings}} 
\authorrunning{Benjam\'in Guti\'errez,  Lo\"ic Peter, Tassilo Klein, Christian Wachinger \thanks{The final publication is available on the MICCAI 17 proceedings}} 
\institute{Department of Child and Adolescent Psychiatry, Psychosomatic and Psychotherapy, Ludwig-Maximilian-University, Munich, Germany
	 \and
Chair for Computer Aided Medical Procedures, Technische Universit\"at M\"unchen, Germany \and
Translational  Imaging Group, University College London, UK \and
SAP SE, Berlin, Germany}

% unless you accept that it will be published
%\institute{Institute\\ Paper 300 }

\toctitle{Lecture Notes in Computer Science}
\tocauthor{Authors' Instructions}
\maketitle

\begin{abstract}
	%The training set is crucial for supervised machine learning. 
	%With the availability of big medical image data, the selection of the training set is becoming more important to 
	With the availability of big medical image data, the  selection of an adequate training set is becoming more important to address the heterogeneity of different datasets.  
	Simply including all the data does not only incur high processing costs but can even harm the prediction. 
	We formulate the smart and efficient selection of a training dataset from big medical image data as a multi-armed bandit problem, solved by Thompson sampling. 
	Our method assumes that image features are not  available at the time of the selection of the samples, and therefore relies only on  meta information associated with the images. Our strategy simultaneously exploits data sources with high chances of yielding useful samples and explores new data regions.	For our evaluation,  we focus on the application of estimating the age from a brain MRI.
	Our results on 7,250 subjects from 10 datasets show that our approach leads to higher accuracy while only requiring a fraction of the training data. 
		
\end{abstract}

\section{Introduction}

%{\red Intro too long, we need to cut it"}

Machine learning has been one of the driving forces for the huge progress in medical imaging analysis over the last years.
%Machine learning has become increasingly important in the field of medical imaging analysis in the last years. 
Of key importance for learning-based techniques is the training dataset that is used for estimating the model parameters. 
Traditionally, medical data has been scarce so usually all available data for a particular task was used for training. Nowadays, many initiatives make data publicly available so that huge amounts of data  can potentially be used to  estimate more accurate models. 
However, just including all the data in the training set is becoming increasingly impractical, since processing the data to create training models can be very time consuming on huge datasets. 
In addition, most processing may be unnecessary because it does not help the model estimation for a given task. 
In this work, we propose a method to select a subset of the data for training that is most relevant for a specific task. 
Foreshadowing some of our results, such a guided selection of a subset of the data for training can lead to a higher performance than using all the available data while requiring only a fraction of the processing time.

The task of selecting a subset of data for training is challenging because at the time of making the decision, we do not yet have processed the data and we do therefore not know how the inclusion of the sample would affect the prediction. However, in many scenarios each image is assigned \emph{metadata}  about the patient (sex, diagnosis,age etc.) or the image acquisition (dataset of origin, location, imaging device, etc.). We hypothetize that some of this information can be useful to guide the selection of samples but it is a priori not clear which information is most relevant and also how it should be distributed. 
To address this problem, we formulate the selection of the samples to be included in a training set as reinforcement learning, 
%Based on this metadata, we design our active acquisition of data as a reinforcement learning framework, 
where a trade-off must be reached between the exploration of new sources of data and the exploitation of sources that have been shown to lead to informative data points in the past. More specifically, we model this as a multi-armed bandit problem solved with Thompson sampling, where each arm of the bandit corresponds to a cluster of samples generated using meta information.

%We propose to formulate the selection of training set as an active learning problem. %the question of selecting the training set as  a multi-arm bandit problem. 
%Each variable from the meta information is modeled as bandit. 
%After selecting an image, it is processed and included in the training data. 
%Depending on whether it improves the prediction accuracy or not, the bandit obtains a reward or not. 
%To address the exploration-exploitation dilemma, as to whether focus on parts of the data that have previously obtained good performance or as to whether explore new areas in the dataset, we propose Thompson sampling. 

In this paper, we apply our sample selection method to  brain age estimation~\cite{franke2010estimating} from MR1 T1 images. 
The estimated age serves as a proxy for biological age, whose difference to the chronological age can be used as indicator of disease~\cite{franke2012brain,gaser2013brainage}. 
The age estimation is a well-suited application for testing our algorithm as it allows us to work with a large number of datasets since the subject's age is one of the few variables that is included in every neuroimaging dataset. 

\subsection{Related Work}

Our work is mostly related to active learning approaches, whose aim is to select samples to be labeled out of a pool of unlabeled data. Examples of active learning approaches applied to medical imaging tasks include the work by Hoi \etal~\cite{hoi2006batch}, where a batch mode active learning approach was presented for selecting medical images for manually labeling the image category. Another active learning approach was proposed for the selection of histopathological slices for manual annotation in~\cite{zhu2014scalable}. The problem was formulated as constrained submodular optimization problem and solved with a greedy algorithm. To select a diverse set of slices, the patient identity was used as meta information. From a methodological point of view, our work relates to the work of Bouneffouf \etal ~\cite{bouneffouf2014contextual}, where an active learning strategy based on contextual multi-armed bandits is proposed.  The main difference between all these active learning approaches and our method is that image features are not available a priori in our application, and therefore can not be used in the sample selection process. Our work also relates to domain adaptation~\cite{pan2010survey}. In instance weighting, the training samples are assigned weights according to the distribution of the labels (class imbalance)~\cite{japkowicz2002class} and the distribution of the observations (covariate shift)~\cite{shimodaira2000improving}. Again these methods are not directly applicable in our scenario because not all the distribution of the metadata is defined on the target dataset.

%In our application, such approaches are applicable for some variables like age and sex, but for other variables like the dataset, they are not applicable. 

\section{Method}

\def\featurematrix{{\mathbf{X}}}
\def\featurevector{{\mathbf{x}}}

\def\meta{{\mathbf{M}}}
\def\trainingfeatures {{\featurematrix_t}}
\def\trainingset{{S_t}}
\def\universalset{{S}}

\def\testingfeatures{{X_s}}

\newcommand{\bbx}{{\bf x}}

\def\targetfeatures{{\features_t}}
\def\sourcefeatures{{\features_s}}
\def\sourcemeta{{\meta_s}}
\def\targetmeta{{\meta_t}}
\def\taskparameters{{\mathbf{p}}}
\def\labelmatrix{{\mathbf{Y}}}
\def\targetlabel{{\prediction_t}}
\def\task{{f}}
\def\real{\mathbb{R}}

\def\integers{\mathbb{Z}}
\def\allsamples{{S}}
\def\sourcesamples{{\allsamples}_s}
\def\targetsamples{{\allsamples}_t}
\def\metavector{\mathbf{m}}
\def\prediction{{y}}
\def\nbandits{{N}}
\def\sample{{s}}
\def\cluster{{C}}
\def\clusterprobability{\mathbf{{\Pi}}}
\def\clusterhiddenprobability{{\hat{\clusterprobability}}}
\def\clusterindex{{i}}
\def\betadistribution{{B}}
\def\reward{{r}}
\def\score{{z}}
\def\nfeatures{{ N_{features} }}
\def\nmeta{{ N_{meta} }}

\def\nparameters{{k}}
\def\totalnumberofsamples{{N_{total}}}
\def\numberoftrainingsamples{{N_{train}}}

\def\featurematrix{{\mathbf{X}}}
\def\featurevector{{\mathbf{x}}}

\def\meta{{\mathbf{M}}}
\def\trainingfeatures {{\featurematrix_t}}
\def\trainingset{{S^T}}
\def\universalset{{S}}

\def\testingfeatures{{X_s}}

\def\targetfeatures{{\features_t}}
\def\sourcefeatures{{\features_s}}
\def\sourcemeta{{\meta_s}}
\def\targetmeta{{\meta_t}}
\def\taskparameters{{\mathbf{p}}}
\def\labelmatrix{{\mathbf{Y}}}
\def\targetlabel{{\prediction_t}}
\def\task{{f}}
\def\real{\mathbb{R}}

\def\integers{\mathbb{Z}}
\def\allsamples{{S}}
\def\sourcesamples{{\allsamples}_s}
\def\targetsamples{{\allsamples}_t}
\def\metavector{\mathbf{m}}
\def\prediction{{y}}
\def\nbandits{{N}}
\def\sample{{s}}
\def\cluster{{C}}
\def\clusterprobability{\mathbf{{\Pi}}}
\def\clusterhiddenprobability{{\hat{\clusterprobability}}}
\def\clusterindex{{i}}
\def\betadistribution{{B}}
\def\reward{{r}}
\def\score{{z}}
\def\nfeatures{{ N_{features} }}
\def\nmeta{{ N_{meta} }}

\def\nparameters{{k}}
\def\totalnumberofsamples{{N_{total}}}
\def\numberoftrainingsamples{{N_{train}}}

\subsection{Incremental Sample Selection}
%Given a a function $\task(\targetfeatures, \taskparameters)\mapsto\labelmatrix$ where $\targetfeatures\in\real^{mxn}$ is a feature matrix where each row corresponds to an observation and each column to a feature, $\taskparameters\in\real^{H}$ is an hyper-parameter vector and $\targetlabel\in\real^{m}$ corresponds to a vector of predictions, our task is that of finding .  
In supervised learning, we model a predictive function   $\task:(\bbx, \taskparameters)\mapsto y$ depending on a parameter vector $\taskparameters$, relating an observation $\mathbf{x}$ to its label $y$. In our application,  $\bbx \in \real^{m}$ is a vector with $m$ quantitative brain measurements from the image and $y \in\real$ is the age of the subject. %, and $\taskparameters$ the parameters of the function that maps brain measurements to a predicted age
%We can define a supervised learning task as that of finding the parameters $\taskparameters$ of a function $\task(\featurematrix, \taskparameters)\mapsto\labelmatrix$ where $\featurematrix\in\real^{mxn}$ is an {\emph observation} matrix, $\taskparameters\in\real^\nparameters$ is a vector of parameters of the learned function and $\labelmatrix\in\real^{mxp}$ is a {\emph prediction} matrix.  
%For example in our case, $\featurematrix$ can be interpreted as brain measurements extracted from an image, $\labelmatrix$ corresponds to the age of the patient and $\taskparameters$ to the parameters of the learning function mapping brain measurements to a predicted age. 
The parameters $\taskparameters$  are estimated by using a training set $\trainingset=\{\sample_1,\sample_2, \ldots, \sample_\numberoftrainingsamples\}$, where each sample  
$\sample = (\featurevector, \prediction )$ is a pair of a feature vector and its associated true label.
Once the parameters are estimated, we can predict the label~$\tilde{y}$ for a new observation~$\tilde{\bbx}$ with $\tilde{y} = f(\tilde{\bbx},\taskparameters^*)$, where the prediction depends on the estimated parameters and therefore the training dataset. 
% by constructing a training matrix $\trainingfeatures$ and its corresponding labels $\labelmatrix_t$ using a training set $\trainingset=\{\sample_1,\sample_2,\sample_\numberoftrainingsamples\}$, where each sample  
%$\sample = \{ \featurevector, \prediction \}$ is a tuple containing a feature vector $\featurevector\in\real^m$ and a label $\prediction\in\real$.

%In a supervised learning scenario, the parameters $\taskparameters$  are approximated by constructing a training matrix $\trainingfeatures$ and its corresponding labels $\labelmatrix_t$ using a training set $\trainingset=\{\sample_1,\sample_2,\sample_\numberoftrainingsamples\}$, where each sample  
%$\sample = \{ \featurevector, \prediction \}$ is a tuple containing a feature vector $\featurevector\in\real^m$ and a label $\prediction\in\real$.The learning procedure consists in finding a parameter vector $\taskparameters^*$  such that  $ \labelmatrix_t \approx \task(\trainingfeatures,\taskparameters^*) \ $. Once the parameters $\taskparameters^*$ are found, they can  be used to obtain the labels $\labelmatrix_s$ of a set of previously unseen observations $\testingfeatures$ by doing $\labelmatrix_s \approx \task(\testingfeatures, \taskparameters^*)$. The estimation of $\testingfeatures$ therefore depends on the samples contained in the training set $\trainingset$ used to learn the parameters $\taskparameters$. 

In our scenario, the samples to be included in the training set $\trainingset$ are selected from a large source set $\universalset = \{h_1,h_2,..,h_\totalnumberofsamples\}$ containing  \emph{hidden} samples of the form $h= \{\hat{\featurevector}, \hat{\prediction}, \metavector  \}$.  
Each $h$ contains hidden features~$\hat{\featurevector}$ and label~$\hat{\prediction}$ that can only be revealed after processing the sample. % by processing the images. 
In addition, each hidden sample possesses a $d$-dimensional vector of metadata $\metavector\in\integers^d$ that encodes characteristics of the patient or the image such as sex, diagnosis, and dataset of origin. 
In contrast to $\hat{\featurevector}$ and $\hat{\prediction}$, $\metavector$ is known a priori and can be observed at no cost.
%
%We target the case where this sample contains some hidden information about its features  and labels, which can not be observed until after the sample is processed.  Additionally, each sample possesses  a vector of meta information $\metavector\in\integers$. The meta information vector encodes some of the characteristics of the patient or the image such as age, gender, and dataset of origin. 
%In contrast to $\featurevector$ and $\prediction$, $\metavector$ is known \emph {a priori} and can be observed at no cost. 
To include a sample $h$ from set $\universalset$ into $\trainingset$, first its features and labels have to be revealed, which comes at a high cost.   
%In our scenario, we consider that the cost of revealing these hidden samples is high. 
Consequently, we would like to find  a sampling strategy that minimizes the cost by selecting only the most relevant samples according to the metadata $\metavector$. 
%In the specific application of age estimation, only the features are hidden and not the label.  %{\red Move somewhere: However there exist applications, like the manual segmentation of images by a rater, where the feature is hidden.}

%, and therefore has to be minimized by including only the most relevant samples in the training set. Since the only information known from the hidden samples $h$ is the meta information given by the vector $\metavector$, we propose a sampling strategy that leverages on this meta information to obtain the most relevant samples. 

\subsection{Multiple Partitions of the Source Data}

In order to guide our sample selection algorithm, we create multiple partitions of the source dataset, where each one considers different information from the metadata $\metavector$.
Considering the $j$-th meta information ($1 \leq j \leq d$), we create the $j$-th partition  $\universalset = \cup_{i=1}^{\eta_j} C_i^j$ with $\eta_j$ a predefined number of bins for $\metavector[j]$. 
As a concrete example, sex could be used for partitioning the data, so $\universalset = C_{\text{female}}^\text{sex} \cup C_{\text{male}}^\text{sex}$ and $\eta_\text{sex} = 2$.
All the clusters generated using different meta information are merged into a set of clusters $\mathcal{C} = \{ C_\iota^j \}$. 
%Our algorithm starts by dividing the source set $\universalset$ into  clusters $\cluster_i \subset \universalset$. 
%Each one of these clusters contains similar samples according to their metadata $\metavector$. 
We hypothesize that given this partitioning, there exist  clusters $C_i \in \mathcal{C}$ that contain more relevant samples than others for a specific task.  Intuitively, we would like to draw samples $h$  from  clusters with a higher probability of returning a relevant sample.  However, since the relationship between the metadata and the task is uncertain, the utility of each cluster for a specific task is unknown beforehand. We will now describe a strategy that simultaneously \emph{explores} the clusters to find out which ones contain more relevant information and \emph{exploits} them by extracting as many samples from relevant clusters as possible.

\def\actionset{{A}}
\def\action{{\mathcal{A}}}
\def\nactions{{N_{actions}}}
\def\timestep{{t}}
\def\reward{{r}}
\def\rewarddistribution{{\Pi}}
\def\rewardprobability{{\Pi}}
\def\metaindex{{k}}
\def\metalabel{{l}}

%\subsection{Sample selection as a multi-armed bandit problem}

%Once we define a partition of our data, we device a strategy to sequentially include new samples to the training dataset $\trainingset$. Our algorithm is based on both exploring which clusters $C_i$ contain the more informative samples and exploiting them by extracting samples from them.  This procedure can be modeled as a Multi-armed bandit problem (MAB) as follows.

 %At each iteration $\timestep$, a new sample is added to the training dataset $\trainingset$. To add a sample, the algorithm decides which cluster $C_i \in \mathcal{C}$ to exploit and randomly draws a training sample without replacement from the cluster  $\sample_\timestep \sim C_i$. After a sample is selected, its corresponding feature vector $\featurevector_\timestep$ and label $\prediction_\timestep$  are added to the training set $\trainingset$. A model $f_t(\mathbf X^t_t,\mathbf{p}_t)$ is trained using $S^T$and then it is used to perform a prediction on a known  set of observations $\tilde y = f(\mathbf{X}^s,\mathbf{p})$.  We then evaluate how useful is sample $\sample_t$ for the given task   yielding a reward $\reward_\timestep\in\real$ based on the improvement of the model $f(\bbx_t,y_t)$ with respect to $f(\bbx_{t-1},y_{t-1})$ .  In our case a reward $\reward_t{=1}$ is given when the $r^2$ score between the prediction $\tilde y_t$ obtained after adding the sample to the training set is higher than the $r^2$ score of  the previous iteration $\tilde y_{t-1}$ or  $\reward_t=-1$ otherwise.   

\subsection{Sample selection as a multi-armed bandit problem}

We model the task of sequential sample selection as a multi-armed bandit  problem. 
%At each iteration $\timestep$, our algorithm must decide to exploit one of the $\nbandits = |\mathcal{C}|$  data clusters. 
At each iteration $\timestep$, a new sample is added to the training dataset $S^T$. 
For adding a sample, the algorithm decides which cluster $C_i \in \mathcal{C}$ to exploit and randomly draws a training sample $\sample_\timestep$ from cluster  $ C_i$.
%Once the cluster is selected, a training sample $\sample_\timestep \sim C_i$ is randomly drawn and 
The corresponding feature vector $\featurevector_\timestep$ and label $\prediction_\timestep$  are revealed and the usefulness of the sample $\sample_t$ for the given task is evaluated, yielding a reward $\reward_\timestep\in\real$. A reward $r_t=1$ is given if adding the sample improves the prediction accuracy of the model and $r_t =-1$ otherwise.

At $t=0$, we do not possess knowledge about the utility of any cluster. However, this knowledge is incrementally built as more and more samples are drawn and their rewards are revealed. 
To this end, each cluster is assigned a distribution of rewards $\rewarddistribution_i$. With every sample the distribution better approximates the true expected reward of the cluster, but every new sample also incurs a cost. 
Therefore, a strategy needs to be designed that explores the distribution for each of the clusters, while at the same time exploiting as often as possible the most rewarding sources.

%At the beginning, we do not have any a priori knowledge about the utility of a cluster. However, we can build this knowledge  incrementally by drawing  samples and observing their rewards. To model the probability of each cluster returning a sample with a positive reward, we assign a probability distribution $\rewarddistribution_i$ that models the expected reward. 
%With every sample the distribution better approximates the true distribution of the expected reward of the cluster, but every new sample also incurs a cost. Therefore, a strategy needs to be designed that explores the distribution for each of the clusters, while at the same time exploiting as often as possible the most rewarding sources. 

%This problem of \emph{exploring} and \emph{exploiting} different actions is also known as the Multi-armed Bandit Problem (MAB). One of the most studied strategies to solve an MAB problem is based on Thompson Sampling (TS)~\cite{thompsonsampling}. 
To solve the problem of selecting from which $C_i$ to sample at every iteration $t$, we follow a strategy based on Thompson sampling~\cite{thompson1933likelihood}  with binary rewards. In this setting, the expected rewards are modeled using a probability $P_i$ following a Bernoulli distribution with parameter $\pi_i \in [0,1]$. We maintain an estimate of the likelihood of each $\pi_i$ given the number of successes $\alpha_i$ and failures $\beta_i$ observed for the cluster $\cluster_i$ so far. 
Successes ($r=1$) and failures ($r=-1$) are defined based on the reward of the current iteration. It can be shown that this likelihood follows the conjugate distribution of a Bernoulli law, i.e., a Beta distribution $Beta(\alpha_i ,\beta_i)$ so that 
\begin{equation}
P(\pi_i | \alpha_i, \beta_i) = \frac{\Gamma(\alpha_i + \beta_i)}{\Gamma(\alpha_i) \Gamma(\beta_i)}(1-\pi_i)^{\beta_i - 1} \pi_i^{\alpha_i - 1}. 
\end{equation}
with the gamma function~$\Gamma$.
At each iteration,  $\hat{\pi}_i$ is drawn from each cluster distribution $P_i$ and the cluster with the maximum  $\hat{\pi}_i$ is chosen. The procedure is summarized in Algorithm~\ref{alg:selection}. 

%The main design choices in our algorithm are the definition of the clusters $\cluster$ and the reward $\reward$.

%
%\begin{algorithm} 
%\caption{Thompson sampling \label{alg:thompson}}
%\begin{algorithmic}[1]
%\For{each  $\timestep=1,2,...$ } 
%\For{each  $\action_i.$ } 
%\State Draw random sample $\theta$ from $\rewarddistribution_i(\beta_\timestep, \alpha_\timestep) $.
%\State Perform the action $\action := \arg \max_i \theta_i(t)$ and obtain a reward $\reward_t$.
%\If {\reward == 0} $\alpha = \alpha +1$ \Else {~$\beta = \beta+1$}
%\EndIf
%\EndFor
%\EndFor
%
%\end{algorithmic}
%\vspace{-.1cm}
%\end{algorithm}
%
%
%\subsection{Modeling Sample selection as a MAB problem}
%
%We will now explain how can we model the sample selection problem as a MAB task, and solve it using the TS algorithm. We  aim at finding the clusters  $\cluster_i$ containing the most relevant samples from our task.  
%Given that we do not know { \emph a priori} which clusters are relevant, the task of selecting a cluster $\cluster$ from which to extract a sample $\sample$ poses an exploration/exploitation dilemma where we want to both discover the more relevant clusters $\cluster$ and extract as much data as possible from them under a limited budget.
%
%In our setting each action $\action$ of the MAB corresponds to extracting a sample $h$ from a cluster $\cluster_\clusterindex$ and revealing its feature $\featurevector$ and label $\prediction$.  If the selected sample $\sample_t$ is useful for the specified task, the bandit gives a reward $\reward$. 

\begin{algorithm} [H]
	\caption{Thompson Sampling for Sample Selection \label{alg:selection}}
	\begin{algorithmic}[1]
		\State $\alpha_i =1, \beta_i =1, \forall i \in \{1, \ldots, N\}$
		\For{$\timestep=1,2,...$ } 
		\For{$i=1, \ldots, N$} 
		\State Draw  $\hat{\pi}_i$ from $Beta(\alpha_i, \beta_i)$.
		\EndFor
		\State Reveal sample $h_t = \{\featurevector_t, \prediction_t, \metavector_t \} $ from cluster $\cluster_j$ where $j := \arg \max_i \hat{\pi}_i$.
		\State Add sample $h_t$ to $\trainingset$ and remove from all clusters. 
		\State Obtain new model parameters $\taskparameters^*$ from updated training set  $\trainingset$.
		\State Compute reward $\reward_t$ based on new prediction~$\tilde{y} = f(\bbx,\taskparameters^*)$.
		\If {$\reward_t == 1$}~$\alpha_j = \alpha_j + 1$ \Else~$\beta_j = \beta_j + 1$
		\EndIf
		\EndFor
	\end{algorithmic}
	\vspace{-.1cm}
\end{algorithm}

\section{Results}

In order to showcase the advantages of our multi-armed bandit sampling algorithm  (MABS), we present an evaluation of our method in the task of estimating the biological age of a subject given a set of volume and thickness features of the brain. We choose this task in particular because of the big number of available brain images in public databases and, the relevance of age estimation as a tool for diagnostic of neuro degenerative diseases~\cite{gaser2013brainage,valizadeh2017age}. For predicting the age, we reconstruct brain scans with FreeSurfer~\cite{FreeSurfer} and extract volume and thickness measurements to create our feature vectors $\featurevector$. Based on these features, we train a regression model for predicting the age of previously unseen subjects. %This approach for age estimation is similar to~\cite{valizadeh2017age} and therefore presents a good setup for evaluating our method.

\subsection{Data}
We work on MRI T1 brain scans from 10 large-scale public datasets: ABIDE~\cite{ABIDE}, ADHD200~\cite{ADHD200}, AIBL~\cite{AIBL}, COBRE~\cite{COBRE}, IXI\footnote{http://brain-development.org/ixi-dataset/}, GSP~\cite{GSP}, HCP~\cite{HCP}, MCIC~\cite{MCIC}, PPMI~\cite{PPMI} and OASIS~\cite{OASIS}.
From all these datasets we obtain a total number of 7,250 images, which is to the best of our knowledge, the largest dataset collected for the task of age prediction.
Since each one of these datasets is targeted towards  different applications, the selected population  is heterogeneous in terms of age, sex, and health status. 
%The distribution of ages among the participants varies from dataset to dataset as well as the meta information obtained from each patient. 
Images are processed with FreeSurfer \cite{FreeSurfer} and thickness and volume measurements extracted. Even though this is a fully automatic tool, the extraction of the feature is a computationally intensive task which is by far the bottleneck of our age prediction regression model. 

%In order to separate the extracted samples into clusters we use the meta information assigned to each image. For our experiments we 
%In theory we could use any  meta information element to create our clusters $\cluster$. %However to maintain clusters of a relevant size we decided to use only age, diagnosis and dataset of origin as the meta information used to generate the clusters. I

\subsection{Age estimation}
We perform age estimation on two different testing  scenarios. In the first, we create a testing dataset by randomly selecting subsets from all the datasets. 
The aim of this experiment is to show that our method is capable of selecting samples that will create a model that can generalize well to a heterogeneous population. In the second scenario, the testing dataset corresponds to a single dataset. In this scenario, we show that the sample selection permits tailoring the training dataset to a specific target dataset. 

\begin{figure}[!ht]
	\centering
	\includegraphics[width=\linewidth]{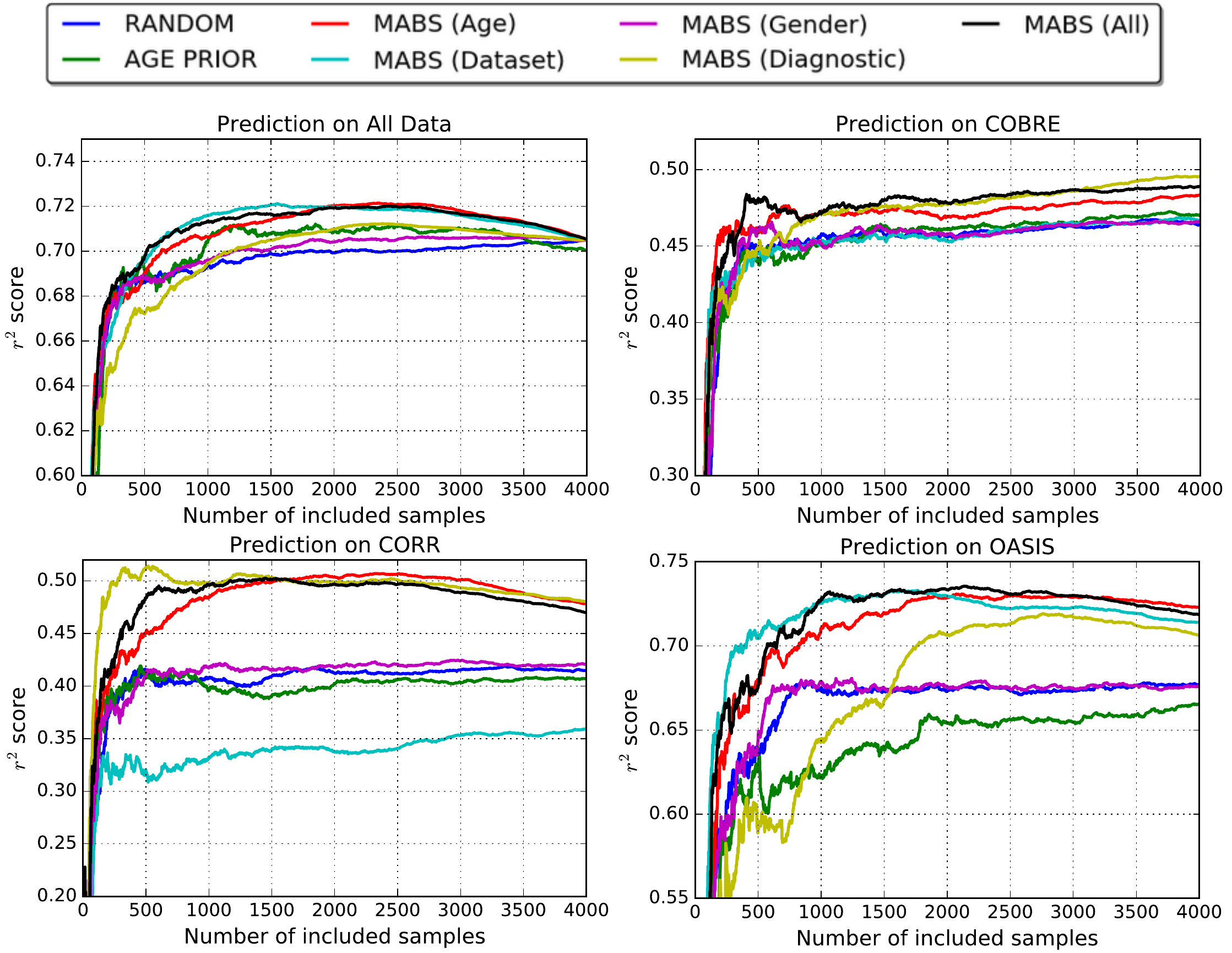}
	\caption{ Results of our age prediction experiments in terms of $r^2$ score.  A comparison is made between MABS using different strategies to build the clusters $\mathcal{C}$, a random selection of samples, and a random selection based on the age distribution of the test data. To improve the visualization of the results, we limit the plot to 4,000 samples.}
	\label{fig:regression_all}
\end{figure}

{\bf Experiment 1.} For the first experiment we take all the images in the dataset and we divide them randomly into three sets: 1) a small validation set of 2\% of all samples  to compute the rewards given to MABS , 2) a large testing set of 48\%  to measure the performance of our age regression task, and 3) a large hidden training set of 50\%, from which samples are taken sequentially using MABS. We perform the sequential sample selection described in Algorithm~\ref{alg:selection} using the following metadata to construct the clusters $\mathcal{C}$:  \emph{age}, \emph{dataset}, \emph{diagnosis}, and  \emph{sex}.
We experiment with considering all of the metadata separately, to investigate the importance of each one, and the joint modeling.  Since we require to evaluate a regression model every time a sample is included, we opted to use ridge regression as our learning algorithm. Rewards $\reward$ are given to  each bandit estimating  and observing if the $r^2$ score of the prediction in the validation set increases. It is important to emphasize that the  testing set is not observed by the bandits in the process of giving rewards.  
%In our experiments we consider clusters constructed using only either \emph{age}, \emph{dataset}, \emph{diagnostic}, or  \emph{gender}. 
%We also test our sample strategy by creating clusters using all meta information at the same time. 
Every experiment is repeated 20 times using different random splits and the mean results are shown. We compare with two  baselines: the first one  (RANDOM) consists of  obtaining samples at random from the hidden set and adding them sequentially to the  training set. As a second baseline (AGE PRIOR), we add samples sequentially by following the age distribution of the testing set .  The results of this first experiment are shown in figure \ref{fig:regression_all} (top left).  In almost all the cases, using MABS as a selection strategy performed better than the baselines. It is important to observe that an increase in performance is obtained not only when the relationships between the metadata and the task are direct, like in the case of the clusters constructed by age, but also when this relationship is not clear like in the case of clustering the images using only dataset or diagnostic information. Another important aspect is that even when the meta information is not informative, like in the case of the clusters generated by sex, the prediction using MABS is not affected. 

{\bf Experiment 2.} For our  second experiment, we perform age estimation with the test data being a specific dataset. This experiment follows the same methodology as the previous one with the important difference of how the datasets are split. This time the split is done by choosing: 1) a small validation set, taken only from the target dataset, 2) a testing set, which corresponds to the remaining samples in the target dataset not included in the validation set, and 3) a hidden dataset containing all the samples from the remaining datasets. The goal of this experiment is to show that our methodology can be deployed when samples have to be selected according to a specific population and prediction task. 
From the results in figure~\ref{fig:regression_all}, we observe that depending on the dataset, using bandits with only one specific source of meta information can actually improve the sample selection algorithm. However, the best meta information for a particular task is different in every case. We  also observe that in general the MABS  using all available meta information extracts  informative samples more efficiently than our baselines and always close to the best performing single meta information MABS. This reinforces our hypothesis that it is hard to define an a priori relationship between the meta data and the task, and is therefore a better strategy to let MABS  select from multiple sources of meta information at once.

\section{Conclusion}
We have proposed a method for efficiently and intelligently sampling a training dataset from a large pool of data. 
The problem was formulated as reinforcement learning, where the training dataset was sequentially built after evaluating a reward function at every step. 
Concretely, we used a multi-armed bandit model that was solved with Thompson sampling. 
The intelligent selection considered metadata of the scan to construct a distribution about the expected reward of a training sample. 
Our results showed that the selective sampling approach leads to higher accuracy than using all the data, while requiring less time for processing the data. 
We demonstrated that our technique can either be used to build a general model or to adapt to a specific target dataset, depending on the composition of the test dataset. 
Since our method does not require to observe the information contained in the images, it can also be applied to predict useful samples even before the images are acquired, guiding the recruitment of subjects.
 
\section{Acknowledgements}
This work was partly funded by SAP SE, the Faculty of Medicine at the University Munich (Förderprogramm für Forschung \& Lehre), and the Bavarian State Ministry of Education, Science and the Arts in the framework of the Centre Digitisation.Bavaria (ZD.B).

\bibliography{biblio-macros,biblio}{}
\bibliographystyle{myplain}

\end{document}